\documentclass[10pt,twocolumn,letterpaper]{article}

\usepackage{cvpr}              

\definecolor{cvprblue}{rgb}{0.21,0.49,0.74}
\usepackage[pagebackref,breaklinks,colorlinks,allcolors=cvprblue]{hyperref}
\usepackage{booktabs}
\usepackage{multirow}
\usepackage[normalem]{ulem}
\useunder{\uline}{\ul}{}
\usepackage[table]{xcolor}
\usepackage{siunitx}

\usepackage{wela}
\usepackage[T1]{fontenc}
\newcommand{\handwriting}[1]{{\wela{\fontseries{bx}\selectfont{#1}}}}
\newcommand{\xhdr}[1]{\noindent\textbf{#1}}

\usepackage{tcolorbox}
\usepackage{xcolor}

\usepackage{algorithm}
\usepackage{algpseudocode}
\usepackage{url}
\newtcolorbox{promptbox}{
    colback=gray!5,      
    colframe=gray!60,    
    boxrule=0.5pt,       
    left=4pt, right=4pt, top=4pt, bottom=4pt,
    arc=2pt,             
    fontupper=\small\ttfamily 
}

\def\paperID{3740} 
\def\confName{CVPR}
\def\confYear{2026}

\newcommand{\name}{\handwriting{Mull-}\texttt{Tokens}} 
\newcommand{\mulltoken}{\handwriting{Mull-}\texttt{Token}} 
\newcommand{\mulltt}{\texttt{<\mull{}\:>}}
\newcommand{\mull}{\handwriting{Mull}}
\newcommand{\ranjay}[1]{}
\newcommand{\guibas}[1]{}
\newcommand{\vincent}[1]{}
\newcommand{\ahmed}[1]{}
\newcommand{\arijit}[1]{}
\newcommand{\chengzhi}[1]{}

\title{\name{}: Modality-Agnostic Latent Thinking}

\author{
    Arijit Ray\textsuperscript{1,4} \quad
    Ahmed Abdelkader\textsuperscript{1} \quad
    Chengzhi Mao\textsuperscript{1} \quad
    Bryan A. Plummer\textsuperscript{4} \quad
    Kate Saenko\textsuperscript{4} \quad \\
    Ranjay Krishna\textsuperscript{2} \quad
    Leonidas Guibas\textsuperscript{1,3} \quad 
    Wen-Sheng Chu\textsuperscript{1} 
    \\[0.7em]
    \textsuperscript{1}\textit{Google} \quad
    \textsuperscript{2}\textit{University of Washington} \quad
    \textsuperscript{3}\textit{Stanford University} \quad
    \textsuperscript{4}\textit{Boston University} \quad
    \\[0.7em] 
    \small{\textbf{\url{https://arijitray.com/multimodal_thinking/}}}
}

\begin{document}
\maketitle
\begin{abstract}

Reasoning goes beyond language; 
the real world requires reasoning about space, time, affordances, and much more that words alone cannot convey. 
Existing multimodal models exploring the potential of \textit{reasoning with images} are brittle and do not scale. 
They rely on calling specialist tools, costly generation of images, or handcrafted datasets.
We offer a simpler alternative -- \name{} -- latent tokens that can be pre-trained to hold intermediate information in either image or text modalities so as to think towards the correct answer. We investigate best practices to train \name{} inspired by latent reasoning frameworks.
We first train \name{} using
supervision from interleaved text-image traces, and then fine-tune without any supervision by only using the final answers.
Across four challenging spatial reasoning benchmarks involving tasks such as solving puzzles and taking different perspectives, we demonstrate that \name{} improve upon several baselines utilizing text-only reasoning or interleaved image-text reasoning, achieving a $+3\%$ average improvement and up to $+16\%$ on a puzzle solving reasoning-heavy split compared to our strongest baseline.
Adding to conversations around challenges in grounding textual and visual reasoning, \name{} offers a simple solution to implicitly think in multiple modalities.

\end{abstract}
\section{Introduction}
\label{sec:intro}



Real-world visual tasks~\citep{vasilyeva2012development,tversky2009thinking} require a synchrony of visual and verbal logical reasoning~\cite{hu2024visual}. 
For example, when solving visual puzzles and IQ tests, people imagine patterns and manipulate visual maps to solve the problem~\citep{mallot2024geometry}. 
Such reasoning—often bundled up as visual/spatial reasoning~\cite{ray2024sat}— is difficult for humans and machines alike~\cite{yang2025_thinking_in_space}. It requires reasoning in space~\cite{lee2025molmoact}, in 3D~\cite{bigverdi2025perception,li2025unfolding}, and often in time~\cite{yang2025_thinking_in_space,brown2025simsvsimulatedinstructiontuningspatial}.
While textual Chain-of-Thought (CoT)~\citep{wei2022chain} has advanced verbal logical reasoning for language models, it falls short on visual reasoning: models still falter when problems demand reasoning about visual causality~\citep{hao2025can,jiang2025corvid}. 

\begin{figure}[t]
\centering
\includegraphics[width=\linewidth]{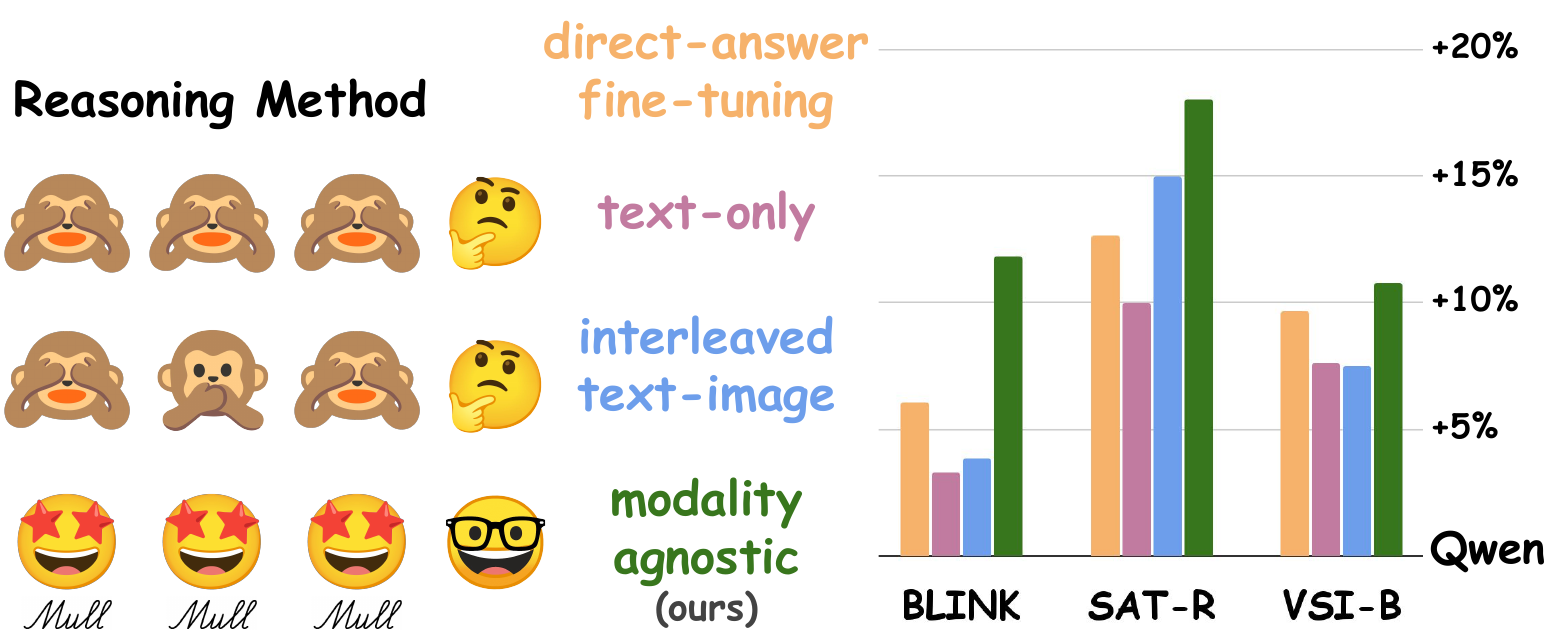}
\caption{Compared to existing approaches for reasoning in text or interleaving image and text, we offer a simpler alternative - modality-agnostic/multimodal thinking in the vision-language space using \name{}.
%
%
%
}
\label{fig:teaser}
\vspace{-2ex}
\end{figure}

To advance visual reasoning, researchers have explored ``think-and-sketch'' strategies~\cite{hu2024visual,openai2025thinking} using interleaved visual thoughts, but with conflicting results~\cite{li2025unfolding}. 
Tool-augmented designs rely on external visual modules~\cite{ma2024tacolearningmultimodalaction} such as cropping tools~\citep{openai2025thinking} or specialized sketching models~\citep{hu2024visual,zhou2024image} - which are incapable of complex visual manipulations and are often brittle~\cite{gu2025thinkmorph}.
Unified models \citep{team2024chameleon,chen2025blip3,chern2024anole} offer a more integrated alternative by generating intermediate images but are expensive to train. 
The latest visual reasoning models instead use image thoughts, either as explicit visual tokens~\cite{bigverdi2025perception} or dense continuous embeddings~\cite{yang2025mirage}. 
However, they require bespoke task-specific training data and hence, do not offer a general recipe for visual reasoning.  
For instance, we find that naively adding \emph{modality-specific} visual thoughts can sometimes hurt: supervising a model to interleave textual thoughts and visual latents actually reduced performance on a visual puzzle task compared to reasoning in text.
Hence, an effective strategy to use the growing availability of multimodal reasoning traces~\cite{li2025zebra,cai2025morse} remains elusive.

To address this gap, we present \name{}: \emph{modality-agnostic} latent tokens that function as a multimodal scratchpad for the model’s internal reasoning, capable of representing \textit{both visual or textual information} as needed. 
During inference, we append these special tokens to the input question, prompting the model to utilize those modality-agnostic slots for arbitrary intermediate computations (spatial mappings, depth predictions, verbal manipulations, etc.) towards the answer, without the need for explicit decoding into text or images.

Inspired by recent latent reasoning frameworks~\cite{pausing_ICLR24,yang2025mirage,yue2025hybrid,shen2025codi}, we develop a two-stage framework for training a model to use \name{} effectively.
The first stage trains the model using multimodal reasoning traces where each \mulltoken{} is \textit{anchored} to a relevant textual or visual concept~\cite{li2025zebra}. For example, one token might learn to encode a visual object layout or a textual symbolic mapping if those are useful for the task at hand.
The second stage relaxes the supervision on \name{} and fine-tunes the model using only the final answer. This stage gives the model an opportunity for optimizing its latent trajectories freely, so as to maximize end-task performance.  \name{} augment a model's vocabulary, similar to special tokens (e.g. <\texttt{plan}> or <\texttt{pause}>) used in reasoning models~\cite{pausing_ICLR24, bigverdi2025perception}. Finally, we follow by a third refinement stage leveraging the latest reinforcement learning (RL) techniques~\cite{guo2025deepseek} to reward good latent chains of ``thinking.''


We fine-tune a Multimodal Language Model (MLM), Qwen2.5-VL (7B)~\cite{bai2025qwen2}, with \name{} as described above. 
We then evaluate the resulting model on a wide range of visual reasoning benchmarks covering both images and videos. 
In particular, we test on spatial reasoning problems such as visual puzzles and IQ-tests from BLINK~\cite{fu2024blink}, VSI-Bench~\cite{yang2025_thinking_in_space} (video spatial reasoning), SAT~\cite{ray2024sat} (action and motion-based spatial relationship reasoning), and MMSI-Bench~\cite{yang2025mmsibenchbenchmarkmultiimagespatial} (multi-image reasoning), offering a thorough evaluation of \name{}' capabilities.

Across all benchmarks, our \name{} significantly improve the model’s reasoning accuracy compared to standard answer-only fine-tuning, textual reasoning, and a recent approach that interleaves explicit text reasoning with image latents \cite{yang2025mirage}.   
On average, \name{} achieve a +3\% absolute accuracy gain over our strongest baseline, which is surprisingly obtained by fine-tuning the base model directly on answers without CoT. The benefits of \name{} are even larger on the most reasoning-intensive tasks, improving +16\% on a visual puzzle split. 

Our results echo the gains reported by prior works when introducing learned visual tokens~\cite{bigverdi2025perception}, but we emphasize that our approach requires far fewer and modality-agnostic tokens. 
Only $10-40$ \name{} are sufficient to yield the improvements above. 
This is a drastic reduction in inference overhead compared to generating hundreds of word tokens in a typical text CoT rationale, or hundreds of visual tokens to represent an image, reminiscent of findings that shorter reasoning traces can be just as effective~\cite{sun2025stop}. 



In summary, our main contribution is an effective reasoning paradigm utilizing modality-agnostic \name{}. Our approach improves upon the best competing approach by $3\%$, with up to $16\%$ on reasoning-heavy splits. \name{} are also faster than producing verbose thoughts~\cite{feng2025videor1} or explicit images as thoughts~\cite{bigverdi2025perception} (20 \mulltt{} vs several hundreds), suggesting a Pareto-dominance of our modality-agnostic approach.
Adjacent to conversations around causal world-model representations for reasoning, \name{} offer a simple approach to reason using visual and textual representations in a unified manner.

\section{Related Work}
\label{sec:rel_works}

The primary theme of our work concerns the incorporation of multi-modal latents to enhance the visual and spatial reasoning capabilities~\cite{chen2025reasoning,ray2024sat,yang2025_thinking_in_space} of MLMs~\cite{zheng2025multimodal, bai2025qwen2}.


\xhdr{Visual Reasoning.}
Building upon powerful perception in vision-language models~\cite{NEURIPS2023_llava,Qwen-VL}, there has been great interest in improving reasoning over both modalities~\cite{zhou2025perception,ke2025explain}.  
Inspired by the success of textual CoT reasoning in language models~\cite{NEURIPS2022_cot_prompting,NEURIPS2023_cot_mystery}, recent works have looked into using similar techniques for visual questions \cite{feng2025videor1,luo2025thinkingdriftsevidentialgrounding}. 
Due to limitations of textual reasoning for visual tasks \cite{gu2025thinkmorph,luo2025thinkingdriftsevidentialgrounding}, many works now reason using visual thoughts as well- but rely on expensive unified models~\cite{team2024chameleon,sun2024emu,wu2024vila,Wu_2025_CVPR,xie2025show,lu2025hyperbagel,zhang2025unified} to explicitly generate images \cite{li2025zebra,gu2025thinkmorph,chern2025thinking,Zhao_2025_CVPR}, or latents~\cite{li2025imagine,yang2025mirage,zhang2024multimodal,bigverdi2025perception,yang2025mirage,zhang2025deepsketcher,chen2025_3Dthinker}, or external tools that may be brittle~\cite{openai2025thinking,hu2024visual}. We aim to incorporate latents as a streamlined and cost-effective alternative without requiring bespoke data \cite{yang2025mirage}. 
While addressing the limitations of textual reasoning using text alone is a valid avenue for exploration \cite{feng2025videor1,li2025think,luo2025thinkingdriftsevidentialgrounding}, we offer a simpler and faster alternative of appending a few modality agnostic thinking tokens. 
Within visual reasoning, there has recently been an emphasis on spatial reasoning~\cite{yang2025_thinking_in_space,zheng2025multimodal,liu2025deconstructing,tong2024cambrian1fullyopenvisioncentric,lee2025molmoact,team2025gemini}
with an increased variety of annotations~\cite{SpatialRGPT_NEURIPS2024,ray2024sat}, benchmarks~\cite{cvbench2025,fu2024blink,team2025gemini,yang2025_thinking_in_space}, and datasets~\cite{feng2025videor1,ray2023colabenchmarkcompositionaltexttoimage,tong2024cambrian1fullyopenvisioncentric,yang2025cambriansspatialsupersensingvideo}.
Our work contributes to training paradigms for improving spatial reasoning~\cite{SpatialRGPT_NEURIPS2024,yang2025cambriansspatialsupersensingvideo,ray2024sat,bigverdi2025perception,daxberger2025mmspatialexploring3dspatial,maninis2025tipstextimagepretrainingspatial,brown2025simsvsimulatedinstructiontuningspatial}.

\xhdr{Latent Reasoning.}
Our thinking tokens are heavily inspired by latent reasoning techniques in language models \cite{pausing_ICLR24,hao2024training,shen2025codi}.
A closely related theme is test-time scaling, allocating more compute at inference time before producing an answer~\cite{snell2024scaling,pausing_ICLR24,kleinman2025e1}.
While some approaches adopt recurrent models over continuous tokens~\cite{hao2024training,geiping2025scaling,zhu2025scaling}, they break standard parallel processing and accumulate errors~\cite{hao2024training,yang2025mirage}. Concurrent works MVoT~\cite{li2025mvot} and MIRAGE~\cite{yang2025mirage} also explore visual thought tokens interleaved with text, similar to the interleaved image-text baseline we compare against. In contrast, our \name{} are modality-agnostic and do not require explicit switching between modalities. We adopt discrete think tokens with a rich internal recurrence achieving substantial gains.

\begin{figure*}[t]
\centering
\includegraphics[width=\linewidth]{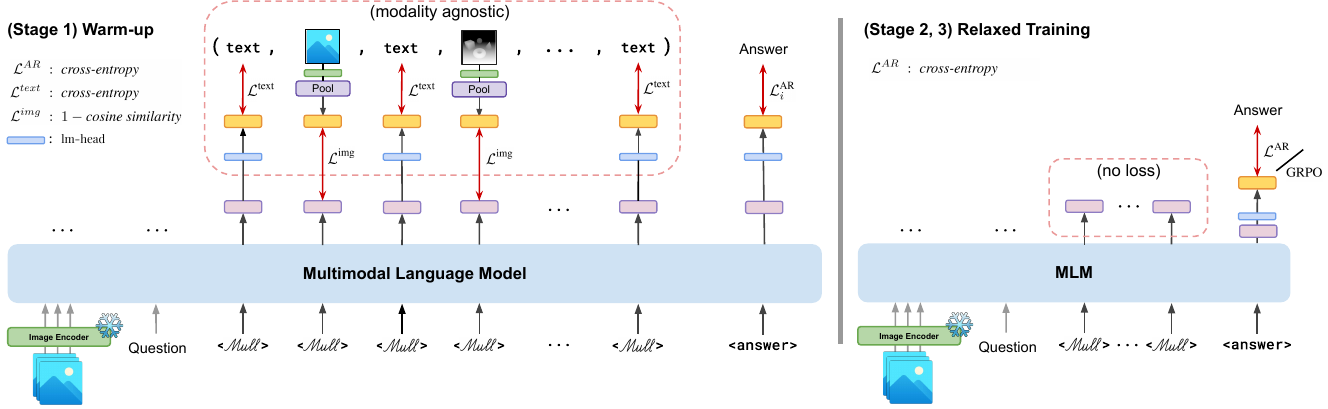}
\caption{
Our \name{} training involves two stages inspired by approaches in latent reasoning. We first pre-train/warm-up our \name{} to hold both image and text modalities depending on the context image/video and query. Next, the model free-form optimizes these \name{} to achieve the final correct answer. We see that pre-training the \name{} with to hold both image and text reasoning traces is key, as opposed to using them simply as extra compute without such pretraining or using text alone.
}
\label{fig:approach}
%
\end{figure*}

\section{Approach}
\label{sec:approach}

Our primary objective is to improve the visual reasoning performance of multimodal language models~\cite{Qwen-VL,NEURIPS2023_llava}.
Formally, the model with parameters $\theta$ implements a conditional distribution
$p_\theta(y | x^{img}, x^{txt})$,
where $x^{\text{img}}$ and $x^{\text{txt}}$ denote the visual and textual inputs, and $y$ is the target answer sequence.

Textual Chain-of-Thought (CoT) augments this input-output mapping with intermediate textual tokens $c^{\text{text}}_{1:T}$ effective for verbal reasoning, but often detrimental for visual tasks~\cite{yang2025_thinking_in_space} due to problems such as drifting from visual inputs~\cite{luo2025thinkingdriftsevidentialgrounding}. 
Recent approaches therefore supervise models to ``think with images,'' \eg, by generating intermediate images or visual sketches $c^{img}_{1:T}$ during reasoning~\cite{hu2024visual,bigverdi2025perception,gu2025thinkmorph}.
However, these methods typically require (i) bespoke multimodal reasoning datasets $\mathcal{D}_{CoT}$ with aligned text–image chains, or (ii) expensive generation of subgoal images that are hard to generalize beyond the specific training domains.

We instead introduce a simpler modality-agnostic mechanism, \name{}, implemented as a fixed-length sequence of special latent tokens $z_{1:K} = (\mulltt{}_1, \dots, \mulltt{}_K)$,
which can carry both image-conditioned and text-conditioned information.
During training and inference, the model can use these tokens purely as \textit{internal compute} to improve visual reasoning, without being constrained to produce textual or image outputs at the intermediate steps.

At a high level, we replace explicit CoT supervision on intermediate text/images with supervision on the \textit{function} of the latent chain: the model is trained so that using $z_{1:K}$ improves $p_\theta(y \mid x^{\text{img}}, x^{\text{txt}})$, while the internal semantics of $z_{1:K}$ are free to adapt.
Below, we formalize our two-stage latent reasoning framework and a third RL refinement stage.

\subsection*{Modality-agnostic \name{}}
Inspired by latent reasoning approaches in NLP~\cite{pausing_ICLR24}, our training is decomposed into three stages (Figure~\ref{fig:approach}):
(1) \textit{warm-up} the \name{} to align with explicit multimodal reasoning steps;
(2) \textit{relax} training to only supervise final answers, encouraging the model to optimize the latent chain internally; and
(3) \textit{refine} the causal utility of latents via GRPO~\cite{guo2025deepseek}.
Our base backbone is Qwen2.5-VL~\cite{bai2025qwen2}, which provides a text decoder and a frozen image encoder.

\subsubsection*{Stage 1: Warm-up the \name{}}
\label{sec:stage1}

Let $\mathcal{D}_{\text{CoT}}$ be a dataset (see Figure~\ref{fig:approach} (left)) of multimodal CoT trajectories.
Each training example consists of an input $(x^{\text{img}}, x^{\text{txt}})$, a reasoning trace
$$c_{1:T} = (c_1, \dots, c_T), \quad 
c_t \in \mathcal{V}^{txt} \cup \mathcal{V}^{img},$$
and a final answer sequence $y_{1:L}$.
Here, $\mathcal{V}^{txt}$ is the text vocabulary and $\mathcal{V}^{\text{img}}$ denotes ``pseudo-tokens'' corresponding to subgoal images (\eg, the jigsaw configuration after inserting a candidate piece).

We construct an interleaved training sequence
$$s = (q_{1:M}, z_1, \tilde{c}_1, z_2, \tilde{c}_2, \dots, z_T, \tilde{c}_T, y_{1:L}),$$
where $q_{1:M}$ encodes the question and context, $\tilde{c}_t$ is the textual substep or an image placeholder for $c_t$, and $z_t = \mulltt{}_t$ is the $t$-th latent token. 
The number of \mulltt{} tokens here matches the length of the reasoning trace in the dataset.
The transformer MLM model is trained in an auto-regressive manner over $s$~\cite{attention_is_all_you_need_NIPS2017}.

Let $h_t^{\text{Mull}} \in \mathbb{R}^d$ denote the hidden state produced by the transformer corresponding to $z_t$.
We apply \textit{two} kinds of supervision:

\begin{itemize}
\item If $c_t \in \mathcal{V}^{txt}$ (a word token), we warp $h_t^{\text{Mull}}$ through the language model head to predict
$p_\theta(c_t \mid s_{<t})$
and minimize the cross-entropy $\mathcal{L}^{text}_t = - \log p_\theta(c_t \mid s_{<t})$.

\item If $c_t \in \mathcal{V}^{img}$ is (a subgoal image $I_t \in \mathbb{R}^{H \times W \times 3}$), we encode it using the frozen Qwen2.5-VL image encoder followed by an average pooling as in~\cite{yang2025mirage}, which we denote by $\bar{g}_{\phi}$: $v_t = \bar{g}_{\phi}(I_t) \in \mathbb{R}^d$.
We then supervise $h_t^\text{Mull}$ via cosine similarity: $\mathcal{L}^{\text{img}}_t = 1 - \cos(h_t^{\text{Mull}}, v_t)$
\end{itemize}

The full Stage-1 objective for a sequence $s$ is
$$\mathcal{L}_{\text{stage1}} 
= \sum_{i \in \mathcal{I}_{\text{AR}}} 
\mathcal{L}^{\text{AR}}_i
+ \lambda_{\text{text}}
\sum_{t \in \mathcal{T}_{\text{text}}} 
\mathcal{L}^{\text{text}}_t
+ \lambda_{\text{img}}
\sum_{t \in \mathcal{T}_{\text{img}}} 
\mathcal{L}^{\text{img}}_t,$$
where $\mathcal{I}_{AR}$ indexes standard question $q_{1:M}$ and answer $y_{1:L}$ tokens, with conventional next-token cross-entropy $\mathcal{L}^{AR}_i$, and $\mathcal{T}^{text}$ / $\mathcal{T}^{img}$ index $\{z_t = \mulltt{}_t\}$ positions followed by text or image steps, respectively.

This setup naturally extends to other modalities (e.g., 3D or trajectory representations) by changing the embedding function $\bar{g}_\phi$ and similarity loss; we leave this extension to future work in the absence of such multimodal CoT data.

\subsubsection*{Stage 2: Relaxed \name{} training}
\label{sec:stage2}

After warm-up, we transition to a relaxed training stage where the latent tokens are no longer explicitly supervised on the next reasoning step.
Let $s' = (q_{1:M}, z_{1:K}, y_{1:L})$ denote the sequence where $z_{1:K}$ is inserted after the question, but the intermediate CoT steps are no longer present.

We now optimize only the answer likelihood and drop all losses on $z_{1:K}$.
$$\mathcal{L}_{\text{stage2}}
= - \sum_{\ell=1}^{L} \log p_\theta(y_\ell \mid s'_{< \ell}),$$
This relaxation is critical: the model is not forced to mimic the potentially suboptimal supervised CoT scaffolding from Stage-1.
Instead, $z_{1:K}$ becomes an internal scratchpad with semantics determined solely by their utility towards minimizing $\mathcal{L}_{\text{stage2}}$ on the final answer.
Note that while $K$ matches the reasoning trace length in Stage-1 (one \mulltt{} per CoT token), at Stage-2 we fix $K$ to a small constant (\eg, $20$), effectively compressing the full reasoning trajectory into a compact latent representation. This compression may also mitigate the reasoning drift observed with verbose text CoT~\cite{luo2025thinkingdriftsevidentialgrounding}, since the model cannot linger on potentially misleading intermediate verbalizations.

A key design choice is the form of recurrence used for multimodal latents.
Let $z^{\text{cont}}_t \in \mathbb{R}^d$ denote continuous recurrent latents as in \cite{hao2024training,yang2025mirage}.
These approaches explicitly iterate over $t$: $z^{\text{cont}}_{t+1} = f_\theta(z^{\text{cont}}_t, x)$.
requiring sequential updates both at training and inference time, and leading to:
(i) \emph{inefficiency}, since they break standard parallel token processing in transformers; and
(ii) \emph{instability}, since errors accumulate over long chains, as empirically noted in~\cite{hao2024training,yang2025mirage}.

Therefore, we adopt discrete token latents as in~\cite{bigverdi2025perception,pausing_ICLR24}.
We allocate a fixed number $K$ of \mulltt{} tokens in the sequence.
Let $H^\text{Mull} = (h_1^\text{Mull}, \dots, h_K^\text{Mull})$ denote their hidden states after the transformer layers; these states are jointly computed via standard self-attention, and then attended to by the answer tokens $y_{1:L}$. We ablate $K$ as a hyperparameter in our experiments.
This design is compatible with transformer parallelism, while still allowing rich internal recurrence through self-attention over $z_{1:K}$.

\begin{figure*}[t]
\centering
\includegraphics[width=0.99\linewidth]{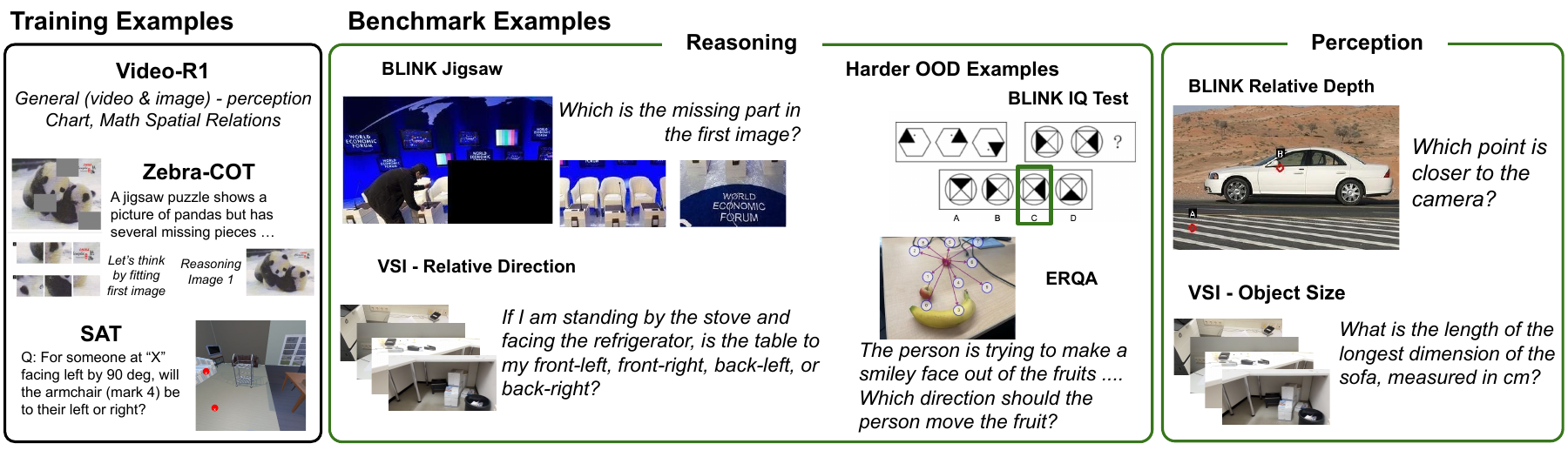}
\caption{Examples of training and our benchmark data. We test on diverse reasoning benchmarks, which includes hard examples that do not have the reasoning outlined in the training sets. Our training consists of two types of data- general video QA's with text reasoning, data of interleaved image text reasoning, and data with only the final answer.}
\label{fig:benchmark_samples}
\end{figure*}

\subsubsection*{Stage 3: RL Refinement with GRPO}
\label{sec:stage3}
Although Stage-2 encourages the model to \textit{use} \name{} to reduce answer loss, it does not enforce that the latent chain is \emph{causally} responsible for the final prediction.
In principle, the model could learn a shortcut mapping from $(x^{\text{img}}, x^{\text{txt}})$ directly to $y$ and ignore $z_{1:K}$, e.g., by placing most of the computation in the final layers conditioned only weakly on the \mulltt{} states.

To explicitly reward latent chains that causally contribute to the correct answer, we introduce a third stage based on GRPO~\cite{guo2025deepseek}, following recent text-based reasoning frameworks~\cite{feng2025videor1,chen2025r1v}.

Let $\pi_\theta(y_{1:L}, z_{1:K} \mid x)$ denote the policy induced by the MLM, where $x = (x^{\text{img}}, x^{\text{txt}})$, and $z_{1:K}$ are internal latents, either sampled or deterministically generated but treated as part of the trajectory.
For each training instance with ground truth answer $y^{*}$, we define a reward
$$r(y_{1:L}, y^{*}) =
\begin{cases}
1, & \text{if } y_{1:L} \text{discrete answers} \\
\mathrm{score}(y_{1:L}, y^{*}) & \text{if numeric / continuous}, 
\end{cases}$$
where $\mathrm{score}$ is a graded similarity, e.g., based on normalized absolute error; exact form is detailed in the supplementary.

In practice, we follow the standard GRPO update that effectively shapes the gradients to emphasize reward-improving trajectories.
Because our \name{} are a fixed discrete prefix of length $K$, and the first answer token $y_1$ is sampled from the final $\mulltt{}_K$ state (\ie, its hidden representation $h_K^{\text{Mull}}$ under self-attention), reward gradients propagate backwards through $h_K^{\text{Mull}}$ primarily and, via self-attention, through the entire latent chain $h_{1:K}^{\text{Mull}}$.
Note that while the policy gradient loss is applied to the sampled answer tokens, the latent tokens are optimized \emph{indirectly}: the answer token logits depend on attending to $h_{1:K}^{\text{Mull}}$, so gradient updates reshape the latent representations.

Thus, GRPO encourages latent trajectories $(h_1^{\text{Mull}}, \dots, h_K^{\text{Mull}})$ that \emph{causally} lead to correct answers, rather than merely co-occurring with them.
Empirically, we find that this third stage further improves performance compared to SFT-only training.

\section{Experiments}
\label{sec:exps}

We evaluate and ablate \name{} on four recent challenging visual reasoning benchmarks 
to assess: i) improvements compared to existing visual reasoning paradigms of reasoning in text or interleaving visual thoughts, ii) the role of \name{}'s multi-modality in those improvements - showing \emph{multimodal} thinking compute is crucial for performance compared to text-only or extra compute without such anchoring, and iii) scaling behavior with latent tokens and recurrence design choices when using \name{} - showing discrete tokens~\cite{pausing_ICLR24} deliver better performance and speed compared to continuous embeddings used in recent work~\cite{yang2025mirage,hao2024training}.


\subsection{Experimental setup}
\label{sec:setup}

\noindent\textbf{Base model.}
We choose a powerful open-source multimodal language model, Qwen2.5-VL (7B) model \cite{bai2025qwen2} as our base since it is also used in existing visual reasoning works~\cite{feng2025videor1,yang2025mirage,luo2025thinkingdriftsevidentialgrounding}. 
All our ablations and baselines are trained using 8 H100 (80GB) GPUs using Deepspeed~\cite{deepspeed_kdd2020}.

\noindent\textbf{Evaluation benchmarks.}
To test performance, we evaluate the answer accuracy on spatial questions for images and videos across four recent benchmarks that test a variety of spatial reasoning capabilities. 
Below we highlight the specific benchmarks we use (also displayed in Figure \ref{fig:benchmark_samples}):

\noindent\textit{BLINK} \cite{fu2024blink} 
We focus on the spatial splits in BLINK. 
We group multi-view reasoning, jigsaw puzzles, and IQ tests as \textit{reasoning} (\emph{Reas} in the Tables) splits since they require comparing/manipulating visual elements.

\begin{table*}[t]
\caption{\name{} (row f, g) improve over direct answer finetuning (row b), and existing methods for visual reasoning tasks using text-only reasoning (row c, d)~\cite{feng2025videor1, chen2025r1v}, or text-image interleaving (row e)~\cite{yang2025mirage}.  Data splits are described in Section~\ref{sec:setup}. 
}
\label{tab:latent_performance}
\small 
\setlength{\tabcolsep}{3.5pt} 
\renewcommand{\arraystretch}{1.2} 
\definecolor{mylightgreen}{HTML}{E8F5E9}
\centering
\begin{tabular}{@{}l ccccccc c cc c r@{}}
\toprule
\textbf{Model} & \multicolumn{7}{c}{\textbf{BLINK}} & \textbf{SAT-R} & \multicolumn{2}{c}{\textbf{VSI}} & \textbf{ERQA} & \textbf{Avg(All)} \\
\cmidrule(lr){2-8} \cmidrule(lr){10-11}
 & \textbf{MV} & \textbf{RelDep} & \textbf{SpRel} & \textbf{Jig} & \textbf{IQT} & \textbf{Avg} & \textbf{Reas} &  & \textbf{Avg} & \textbf{Reas} &  &  \\ \midrule
a. Qwen2.5-VL (7B)  & 39.00 & 61.29 & \textbf{92.38} & 58.66 & 25.33 & 55.33 & 41.00 & 59.00 & 23.96 & 22.96 & \textbf{38.91} & 44.30 \\ \midrule[0.8pt]
b. + DirAns FT & 57.14 & \textbf{87.09} & 74.12 & 58.66 & 30.00 & 61.40 & 48.60 & 71.66 & 32.40 & 30.65 & 38.00 & 50.87 \\
\cmidrule(lr){1-13}
c. + TextCoT FT & 53.38 & 74.19 & 67.83 & 69.30 & 25.33 & 58.01 & 49.34 & 68.33 & 30.47 & 31.04 & 38.78 & 48.90 \\
\rowcolor{mylightgreen} \quad \textit{$\Delta$ (vs. DirAns)} & \textit{-3.76} & \textit{-12.90} & \textit{-6.29} & \textit{+10.64} & \textit{-4.67} & \textit{-3.39} & \textit{+0.74} & \textit{-3.33} & \textit{-1.93} & \textit{+0.39} & \textit{+0.78} & \textit{-1.97} \\
d. \quad + GRPO & 54.88 & 71.77 & 69.23 & 72.00 & 25.33 & 58.64 & 50.74 & 69.00 & 30.34 & 30.15 & 36.00 & 48.50 \\
\rowcolor{mylightgreen} \quad \textit{$\Delta$ (vs. DirAns)} & \textit{-2.26} & \textit{-15.32} & \textit{-4.89} & \textit{+13.34} & \textit{-4.67} & \textit{-2.76} & \textit{+2.14} & \textit{-2.66} & \textit{-2.06} & \textit{-0.50} & \textit{-2.00} & \textit{-2.37} \\ \cmidrule(lr){1-13}
e. + Interleave Im-Txt & 57.14 & 69.36 & 75.52 & 68.67 & 25.33 & 59.20 & 50.38 & 74.00 & 30.27 & 32.96 & 38.50 & 50.49 \\
\rowcolor{mylightgreen} \quad \textit{$\Delta$ (vs. DirAns)} & \textit{+0.00} & \textit{-17.73} & \textit{+1.40} & \textit{+10.01} & \textit{-4.67} & \textit{-2.20} & \textit{+1.78} & \textit{+2.34} & \textit{-2.13} & \textit{+2.31} & \textit{+0.50} & \textit{-0.38} \\ \cmidrule(lr){1-13}
f. + \name{} & 63.15 & 83.06 & 81.81 & 74.00 & \textbf{32.00} & 66.80 & 56.38 & \textbf{77.66} & 32.98 & 32.85 & 38.25 & 53.92 \\
\rowcolor{mylightgreen} \quad \textit{$\Delta$ (vs. DirAns)} & \textit{+6.01} & \textit{-4.03} & \textit{+7.69} & \textit{+15.34} & \textit{+2.00} & \textit{+5.40} & \textit{+7.78} & \textit{+6.00} & \textit{+0.58} & \textit{+2.20} & \textit{+0.25} & \textit{+3.05} \\
g. \quad + GRPO & \textbf{64.66} & 83.87 & 81.81 & \textbf{74.67} & 30.66 & \textbf{67.13} & \textbf{56.66} & 77.00 & \textbf{33.51} & \textbf{33.49} & 38.50 & \textbf{54.04} \\
\rowcolor{mylightgreen} \quad \textit{$\Delta$ (vs. DirAns)} & \textit{+7.52} & \textit{-3.22} & \textit{+7.69} & \textit{+16.01} & \textit{+0.66} & \textit{+5.73} & \textit{+8.06} & \textit{+5.34} & \textit{+1.11} & \textit{+2.84} & \textit{+0.50} & \textit{+3.17} \\ \bottomrule
\end{tabular}
\end{table*}
    
\noindent\textit{SAT-Real (SAT-R)} \cite{ray2024sat}    
This includes questions that require reasoning about actions based on one/two image(s). 

\noindent\textit{VSI-Bench (VSI-B)} \cite{yang2025_thinking_in_space}
This is a video benchmark with questions about relationships and distances from other perspectives, and optimal routes.  
We group relative direction, relative distance, and route planning into the reasoning splits since they involve perspective-taking or planning.
Absolute numbers may differ from some concurrent works due to differences in frame count, answer matching logic, and system prompts. We use consistent hyperparameters across all ablations.  We also include some broader datasets to test generalization, like ERQA~\cite{team2025gemini}, MMSI-Bench~\cite{yang2025mmsibenchbenchmarkmultiimagespatial}, and SiteBench~\cite{Wang_2025_ICCV}. 


\begin{table}[t]
\centering
\caption{Warming up \name{} with multimodal image-text CoT (MM warm-up, row e) is crucial for performance, compared to no warm-up (row c), text-only warm up (row d). This shows that \mulltt{} are not simply extra computation, the multimodal reasoning information in them is crucial. Data splits are described in Section~\ref{sec:setup}.}
\label{tab:latent_warmup}
\definecolor{mylightgreen}{HTML}{E8F5E9}
\setlength{\tabcolsep}{3pt}
\renewcommand{\arraystretch}{1.15}
\footnotesize
\begin{tabular}{@{}l cc cc c c@{}}
\toprule
\textbf{Model} & \multicolumn{2}{c}{\textbf{BLINK}} & \multicolumn{2}{c}{\textbf{VSI}} & \textbf{SAT-R} & \textbf{ERQA} \\
\cmidrule(lr){2-3} \cmidrule(lr){4-5}
& Avg & Reas & Avg & Reas & & \\ \midrule
a. QwenVL & 55.3 & 41.0 & 24.0 & 23.0 & 59.0 & \textbf{38.9} \\
b. + DirAns FT & 61.4 & 48.6 & 32.4 & 30.7 & 71.7 & 38.0 \\ \midrule[0.8pt]
c. + No warm-up & 59.2 & 45.2 & 29.8 & 27.4 & 67.3 & 37.8 \\
d. + Txt warm-up & 65.9 & 52.9 & 32.0 & \textbf{33.5} & 71.3 & 38.5 \\
e. + MM warm-up & \textbf{66.8} & \textbf{56.4} & \textbf{33.0} & 32.9 & \textbf{77.7} & 38.3 \\
\bottomrule
\end{tabular}
\end{table}

\noindent\textbf{Training datasets.}
\label{sec:train_datasets}
We use three datasets: Video-R1, Zebra-CoT, and SAT as shown in Figure \ref{fig:benchmark_samples}. 

\noindent\textit{Video-R1}~\cite{feng2025videor1} is a general video reasoning dataset with reasoning traces containing 160K examples. 
It contains videos and images with questions spanning multiple topics like chart understanding, spatial reasoning, and perception tasks. 
When we use the dataset without the reasoning trace and supervise the answer directly, we denote that as Video-R1-Ans  in the tables.

\noindent\textit{Zebra-CoT}~\cite{li2025zebra} is a recent image-text interleaved CoT dataset with reasoning traces that include both text and sub-goal images as reasoning traces towards the answer for a range of visual and spatial tasks such as playing chess, solving puzzles, and counting. 

\noindent\textit{SAT}~\cite{ray2024sat} contains 175K simulated examples of spatial reasoning questions and answers with both single-image and multi-image inputs. 


\noindent\textbf{Baselines.}
All baselines are trained using the training datasets listed above.
We explore four baseline approaches. 

\noindent\textbf{\small{+DirAns}} trains the base model directly on the final answer without any reasoning traces. 

\noindent\textbf{\small{+TextCoT}} trains using the direct answers from SAT~\cite{ray2024sat} and ZebraCoT~\cite{li2025zebra} as well as the text-based reasoning traces from Video-R1~\cite{feng2025videor1}.
\noindent\textit{\small{+GRPO}} further fine-tunes the above with GRPO~\cite{chen2025r1v,guo2025deepseek,feng2025videor1} to refine the reasoning trace using only the final answer. We use a group size of 2 following recent work~\cite{geng2025deltalearninghypothesispreference} and a batch size of $256$ with gradient accumulation for stability.  

\noindent\textbf{\small{+Interleave Im-Txt}} replicates a recent approach~\cite{yang2025mirage} that uses latents to encode visual information and \textit{retains} explicit, text-based reasoning traces. 
We follow a similar two-stage approach to our method. Instead of warming up the \mulltt{} with both image and text, we interleave text with image-only supervision for \mulltt{}. 
During stage 2, we relax the loss on the image \mulltt{} tokens. 

\subsection{Main results}

\textbf{Text-based reasoning improves base model, but hurts compared to direct answer fine-tuning.}
We observe that supervising the model with text-based reasoning traces, while improving upon the base model performances as also found in concurrent work \cite{feng2025videor1}, \textit{hurts} performance compared to directly tuning on the final answer of the same datasets.
The results in Table \ref{tab:latent_performance} show that direct answer fine-tuning (row b) is a strong baseline, outperforming the text-based reasoning fine-tuning (row c) and also after GRPO optimization of the reasoning chains (row d).  
However, we do note that text-based reasoning tends to help on more reasoning-heavy splits compared to just fine-tuning with the final answer - BLINK Jigsaw and reasoning average (by $\sim 2.1\%$) and VSI-Bench reasoning ($\sim 0.4\%$).

\noindent \textbf{Interleaving image thoughts fails to improve significantly.}
Recall that the intuitive solution to ``fix'' text-based reasoning is to include visual thoughts (image latents) along with explicit text thoughts~\cite{yang2025mirage}. 
The results of such an approach applied to our datasets (described in Section~\ref{sec:setup}) are shown in Table \ref{tab:latent_performance} in row e. 
We see that it indeed performs better than text-based reasoning (row e vs d), with more gains seen on video reasoning splits (VSI reasoning) and on splits that require reasoning about action consequence and perspectives (SAT-Real and ERQA). 
However, it fails to improve overall from direct-answer finetuning across all benchmarks except SAT~\cite{ray2024sat}.
It also fails to improve on reasoning splits that are harder and out-of-domain such as BLINK IQ Tests.
Empirically analyzing the interleaved reasoning outputs reveal two weaknesses - i) the model \emph{rarely} switches to image thoughts and resorts to mostly text reasoning and hence, suffers the same weaknesses, and ii) if we prompt the model to force it to switch to image thoughts (more details in the supplementary), it actually reduces performance - \eg on BLINK, performance went down by $2\%$ when forced to reason with image thoughts. We observe that the text still suffers from drifting away from visual inputs even after a visual thought (examples in supplementary). 
This suggests that while interleaving image thoughts may be helpful sometimes, the model still cannot effectively use it to think towards an accurate answer.

\noindent\textbf{Using modality-agnostic \name{} performs best overall.}
As shown in Table \ref{tab:latent_performance} row f and g, we see that our \name{}, which are agnostic to text or image modality, perform best overall, including reasoning splits that are harder out-of-domain tasks.
Our \name{} (after stage 2) (row f) improves compared to direct-answer finetuning (row b), where various other reasoning approaches (row c, d, e, f) fail to. 
Notably, we have strong improvements in reasoning about camera movements, jigsaw puzzles and IQ tests (under BLINK column) that require reasoning with visual and symbolic modalities.  
Finally, we observe that latest advancements in RL, GRPO refinement, when applied to our \name{} (row g) can further improve the reasoning-heavy splits in BLINK and VSI-Bench.
Improvements to interleaved image-text reasoning (row e) suggest that effective thinking compute requires \name{} to be \emph{modality-agnostic} rather than restricted to vision.
Notably, \name{} achieve these gains using only $20$ tokens, compared to $200$--$500$ tokens in a typical text CoT rationale, translating to significant inference speedup over verbose reasoning approaches.
We also note that on ERQA, all fine-tuned variants perform close to the base model. We attribute this to ERQA being perception-heavy: many questions ask about directly visible object states or trajectories, limiting the headroom for reasoning-based improvements.

\noindent\textbf{\name{} generalize to more complex multi-image benchmarks}
To test broader generalization, we evaluate on MMSI-Bench~\cite{yang2025mmsibenchbenchmarkmultiimagespatial} (multi-image spatial) and SiteBench~\cite{Wang_2025_ICCV} (general spatial, 3K randomly sampled due to compute constraints). Table~\ref{tab:generalization} shows we improve multi-step reasoning (MSR, +1.2\%), judging attributes from different perspectives (Appr. +8.0\%), and also general spatial tasks (SiteBench +2.1\%), suggesting broader generalization.

\begin{table}[t]
\caption{Performance generalization on other multi-image or general-purpose spatial benchmarks.}
\label{tab:generalization}
\centering
\small
\setlength{\tabcolsep}{3pt}
\begin{tabular}{@{}lcccc@{}}\toprule
& \multicolumn{3}{c}{\textbf{MMSI-Bench}} & \textbf{SITE-B} \\
\cmidrule(lr){2-4}
\textbf{Model} & \textbf{MSR} & \textbf{Attr.} & \textbf{Avg} & \textbf{Avg} \\
 \midrule
Qwen 7B & 25.8 & 18.2/25.0 & 25.9 & 61.3 \\
+DirAns & 27.4 & 25.3/25.7 & 28.8 & 63.5 \\
+\name{} & \textbf{28.7} & \textbf{33.3/28.6} & \textbf{29.0} & \textbf{65.6} \\ \bottomrule
\end{tabular}
\vspace{-2mm}
\end{table}

\noindent\textbf{Adding \name{} can improve answering accuracy while also predicting the text rationale.}
While \name{} without text-based reasoning improves overall performance, predicting the text-based reasoning (or rationale) can often be of educational and interpretable value.
Hence, we explore if \name{} can also be appended before predicting both the text rationale and the final answer.
Intuitively, this allows the model to think in an abstract modality-agnostic space before verbalizing the rationale and predicting the answer.
To perform this analysis, we include both the text rationale and the answers from the same training datasets in Stage 2 (Section~\ref{sec:stage2}) of our training after the \texttt{Mull} tokens.
Since we do not have a reliable way to evaluate the rationales, we report the final answer accuracy in Table~\ref{tab:mull_w_expl} and show some qualitative results in Figure~\ref{fig:mull_qual_wtxt}. 
Interestingly, we note that the model decides to use text reasoning for certain tasks and just the \name{} tokens for tasks that may not need text reasoning to answer accurately. 
We observe that the answering accuracy after predicting the text rationale (row d) is higher than if we were to directly predict the rationale and answer (row b).
While \name{} are not directly decodable into human-readable output, they can be paired with explicit text rationales, retaining the accuracy benefits of latent reasoning \emph{while also} producing interpretable reasoning traces.

\begin{table}[t]
\caption{We can add \name{} before predicting the explicit text reasoning and the final answer (row d). This improves answering performance compared to directly predicting the text reasoning (row b) or having interleaved visual thoughts with text reasoning (row c) before predicting the answer.}
\label{tab:mull_w_expl}
\centering
\footnotesize
\setlength{\tabcolsep}{3pt}
\begin{tabular}{@{}lccccc@{}}\toprule
\textbf{Model}           & \textbf{BLINK}  & \textbf{SAT-R} & \textbf{VSI}  & \textbf{ERQA}   & \textbf{Avg}    \\ \midrule
a. Qwen 2.5 VL 7B           & 55.3          & 59.0           & 24.0          & \textbf{38.9} & 44.3          \\
b. Text               & 58.0          & 68.3         & 30.5          & 38.8          & 48.9          \\
c. Img + Text & 59.2          & \textbf{74.0}  & 30.3          & 38.5           & 50.5          \\
d. \name{} + Text   & \textbf{62.4} & 72.0         & \textbf{32.6} & 37.3          & \textbf{51.1} \\ \bottomrule
\end{tabular}

\end{table}

\begin{figure*}[t]
\centering
\includegraphics[width=0.95\linewidth]{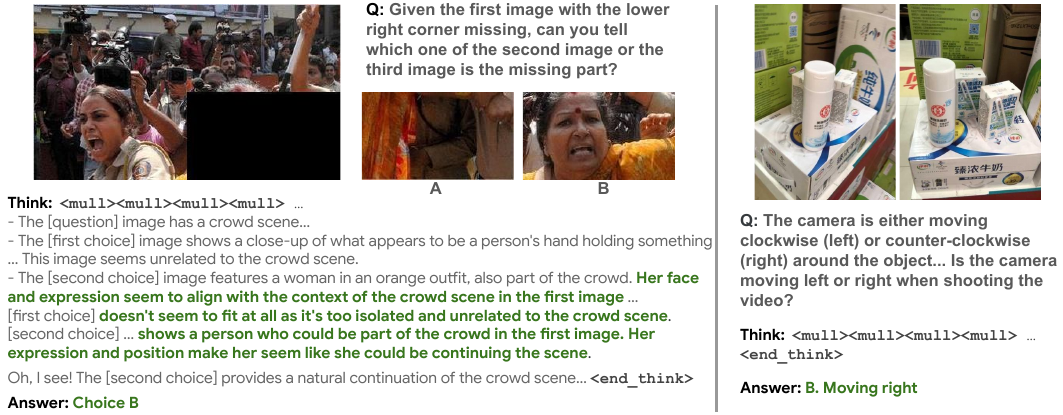}
\caption{
\name{} tokens can be used in conjunction with text reasoning and the model can effectively decide when to also avoid using text reasoning depending on the task. For the example on the left, the model accurately uses \name{} along with some textual descriptive cues to reason about the missing piece. For the example on the right, the model decides to simply use the \name{} to directly predict the answer since textual descriptions are likely not helpful to detect camera motion.
}
\label{fig:mull_qual_wtxt}
\end{figure*}

\subsection{Ablations}
\label{sec:ablations}

We further ablate what causes the performance gains. 
We explore whether the \name{} utilize the multimodal information in its reasoning latents. The model could just be taking advantage of a wider computational pathway of meaningless tokens instead.

\xhdr{No Warm-up.}
If the model were to simply use \name{} as a wider computational pathway, we would see gains if we only fine-tuned according to Stage 2 (Section~\ref{sec:stage2} with the new tokens (without any multimodal information imparted into them) as extra compute for the model to optimize to arrive at the final answer. Hence, we perform this ablation and perform only Stage 2 tuning using direct answers from our training datasets 

\xhdr{Text-only Warm-up.}
To ablate whether multimodal - \ie including images with the text - is necessary, or text alone is sufficient, we warm-up the \name{} using \textit{only} the text-based CoT traces from our training data, discarding any traces that include interleaved images. We still use the final answers from the image COT trace dataset to not let domain differences in the datasets affect the performance. 

\xhdr{Our final approach.}
Our final \name{} approach, warms up the tokens on all available interleaved image-and-text reasoning traces in stage 1 (denoted by \textbf{\small{MM warm-up}}) before performing stage 2 training without any constraints on our latent tokens and only using the final answers from all our training datasets. 

\xhdr{Continuous Embeddings vs Discrete Tokens}
Recall that the choices of using latent \name{} tokens in stage 2 of the training (Section \ref{sec:stage2} and during inference are to use continuous embeddings that are recurrently fed in for $k$ steps~\cite{hao2024training}, or use discrete \mulltt{} tokens~\cite{pausing_ICLR24}. For the continuous ablation, in stage 2, for each latent \texttt{MULL}$_k$, we feed in the previous \texttt{MULL}$_{k-1}$ embedding output before the language head of the Qwen model - both during inference and training.

\begin{figure}[t]
\centering
\includegraphics[width=\linewidth]{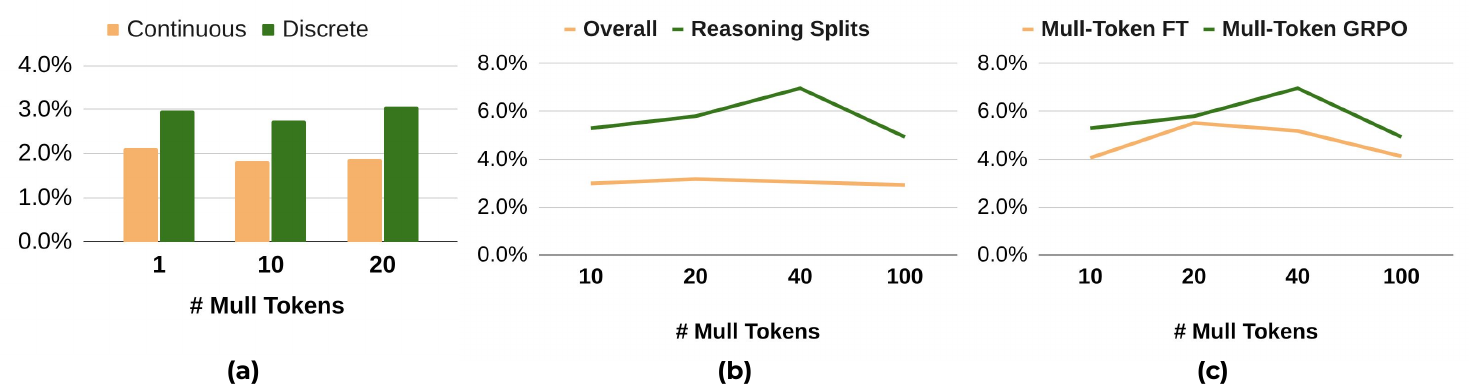}
\caption{
Hyperparameter choices in latent design. (a) Choice of continuous embeddings vs discrete tokens: We see that discrete performs better for each $k$ (number of latent at test time) values. (b) The impact of number of latent tokens at test time: we see that too many latents can hurt performance. Reasoning splits tend to improve with more \name{}, but too many is still detrimental. (c) We see that performance scales with number of latent tokens more after GRPO, since GRPO rewards the latent chain causally.}
\label{fig:num_tokens}
\end{figure}

\subsection{Analysis}

\noindent \textbf{\name{} with both image and text anchoring is crucial to performance.}
While \name{} is effective, a question remains whether \name{} represents reasoning in a multimodal space or simply using extra computation~\cite{pausing_ICLR24} to improve performance.  
Hence, we ablate the choice of warming-up our \name{} as described in Section~\ref{sec:ablations} and present the results in Table \ref{tab:latent_warmup}.
We find that multimodal warm-up allowing the \name{} to be multimodal - \ie can hold meanings in either text or images - to be beneficial for performance. 
\name{} used simply as extra compute (row c) improves upon the base model by $4.2\%$, however, undeperforms the direct answer baseline (row b)
\name{} with text-only warm-up (row d) offers only a marginal 1.07\% improvement over the direct answer fine-tuning baseline (row b), showing that a text-only warm-up, while somewhat beneficial, is still insufficient.
Finally, warming up with both image and text achieves the highest gains ($+3.05\%$) over the direct answer fine-tuning. 
This suggests that for \name{} to be effective, they \textit{must} be pre-trained to handle \textit{both} visual and textual information simultaneously.

\label{sec:finding4}
\noindent \textbf{Discrete latent tokens over continuous embeddings.}   
Both strategies improve upon the direct answer finetuning baseline as shown in Figure \ref{fig:num_tokens} - plot (a).
However, discrete tokens perform better. Performance also starts to degrade with more continuous latents since errors accumulate with longer continuous embeddings, also seen by concurrent works \cite{yang2025mirage}. 
Discrete also allows us to use the latest optimization tricks, such as token parallelism in standard frameworks for MLMs, resulting in significantly faster training and inference.  

\label{sec:finding5} 
\noindent \textbf{Scaling effect of latent tokens.} We investigate the effect of scaling the number of latent tokens used during inference - both after the \name{} FT and GRPO stages. Our model was trained using 20 latents.
The results in Figure \ref{fig:num_tokens} - (b) and (c) show the tradeoff.
We first see that reasoning tasks tend to benefit more with a higher number of tokens as shown in plot (b), but performance degrades with too many tokens- reminiscent of degradation due to overthinking in LLMs~\cite{wei2025stopspinningwheelsmitigating}. 
We see that performance scales more positively after the GRPO phase, as shown in (c). 
We believe this might be because GRPO rewards the optimal \mulltt{} chains on-policy, unlike SFT.

\section{Discussion}

\xhdr{Limitations.}
Our experiments focus on the Qwen2.5-VL (7B) model~\cite{bai2025qwen2}; while effective, the generalization of our findings to other backbones and model scales remains to be verified. Based on trends in latent reasoning for language models~\cite{pausing_ICLR24,shen2025codi}, where gains grow with model capacity, we anticipate our approach to benefit similarly from larger base models.
A more intrinsic limitation is the lack of direct interpretability for the latent tokens. As these tokens are modality-agnostic—representing neither pure text nor pure image features—they cannot be directly decoded into human-readable output.
However, from a performance angle, a few \name{} improve performance over verbose, albeit interpretable, text thoughts.
As shown in Table~\ref{tab:mull_w_expl}, \name{} can also be used \emph{in conjunction} with interpretable text thoughts, offering both accuracy gains and interpretability. 

\xhdr{Future work.}
We identify several key avenues for future work. First, extension of our approach to include diverse modalities such as 3D point clouds, audio, and other structured data, a direction currently constrained by the scarcity of suitable reasoning datasets. 
Finally, we see significant potential in integrating learning from world models~\cite{genie3} to discover robust causal reasoning chains, which is essential for bridging the remaining performance gap between current models and human capabilities.

\xhdr{Conclusion.} In this work, we present a novel fine-tuning paradigm that significantly improves multimodal reasoning compared to existing text-only or image-text interleaved reasoning approaches. Our modality-agnostic latent \mulltt{} tokens are Pareto-dominant: they improve performance (+$3\%$) while being computationally more efficient (20 tokens vs 100's). We hope this work paves the way for efficiently incorporating causal multimodal reasoning in autoregressive AI models.




\subsection*{Acknowledgments}
We wish to thank Howard Zhou, Andre Araujo, Kevis-Kokitsi Maninis, and Ye Xia for helpful initial discussions on the directions. We also thank Mi Luo, Ellis Brown, and Saining Xie for their thoughts that helped polish the paper. We would like to thank the Google XCloud and Cloud Storage Teams for providing a solid infrastructure to train and evaluate open-source models on open-source datasets and benchmarks. Finally, we would also like to acknowledge the excellent codebases provided by Video-R1 and MIRAGE authors that we built upon. RK was partially supported by SRC Jump 2.0 CoCoSys.  

{
    \small
    \bibliographystyle{ieeenat_fullname}
    \bibliography{main,extra}
}



\label{sec:suppl}

\section{Appendix}

\label{sec:suppl}

In this supplementary document, we include further ablations in the training design choices, more details, and more insights into the shortcomings of related existing work versus our approach using qualitative examples. 
Finally, we also provide some qualitative examples to demonstrate our insights. 


\definecolor{cotcolor}{RGB}{230, 240, 255}   
\definecolor{latentcolor}{RGB}{240, 255, 240} 
\definecolor{sftcolor}{RGB}{255, 245, 235}    

\begin{figure*}[t]
\centering
\includegraphics[width=0.9\linewidth]{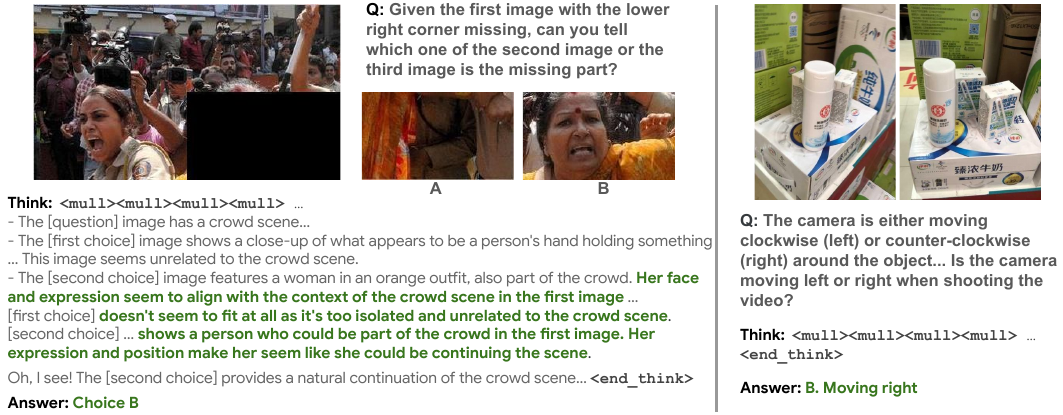}
\caption{
\name{} tokens can be used in conjunction with text reasoning and the model can effectively decide when to also avoid using text reasoning depending on the task. For the example on the left, the model accurately uses \name{} along with some textual descriptive cues to reason about the missing piece. For the example on the right, the model decides to simply use the \name{} to directly predict the answer since textual descriptions are likely not helpful to detect camera motion.
}
\label{fig:mull_qual_wtxt}
\end{figure*}

\begin{table*}[t]
\centering
\caption{Performance with different prompt strategies interleaving image and text thoughts. Explicitly instructing the model to think with images hurts performance compared to letting the model choose. }
\label{tab:im_text_interleave}
\small
\setlength{\tabcolsep}{4.5pt}
\begin{tabular}{@{}lcccccccccccc@{}}
\toprule
\multirow{2}{*}{\textbf{Model}} & \multicolumn{7}{c}{\textbf{BLINK}} & \multirow{2}{*}{\textbf{SAT-R}} & \multicolumn{2}{c}{\textbf{VSI}} & \multirow{2}{*}{\textbf{ERQA}} & \multirow{2}{*}{\textbf{All Avg}} \\ \cmidrule(lr){2-8} \cmidrule(lr){10-11}
 & \textbf{MV} & \textbf{RelDep} & \textbf{SpRel} & \textbf{Jigsaw} & \textbf{IQT} & \textbf{Avg} & \textbf{Reason} &  & \textbf{Avg} & \textbf{Reason} &  &  \\ \midrule
Qwen 2.5 VL 7B & 39.0 & 61.3 & 92.4 & 58.7 & 25.3 & 55.3 & 41.0 & 59.0 & 24.0 & 23.0 & 38.9 & 44.3 \\
+ Img-Txt (free-form) & 57.1 & 69.4 & 75.5 & 68.7 & 25.3 & 59.2 & 50.4 & 74.0 & 30.3 & 33.0 & 38.5 & 50.5 \\
+ Img-Txt (force im thought) & 49.6 & 73.4 & 73.4 & 69.3 & 24.7 & 58.1 & 47.9 & 65.0 & 27.2 & 30.7 & 39.0 & 47.3 \\
+ \name{} (our) & 63.1 & 83.1 & 81.8 & 74.0 & 32.0 & 66.8 & 56.4 & 77.7 & 33.0 & 32.9 & 38.2 & 53.9 \\ \bottomrule
\end{tabular}
\end{table*}

\begin{figure*}[ht]
\centering
\includegraphics[width=0.8\linewidth]{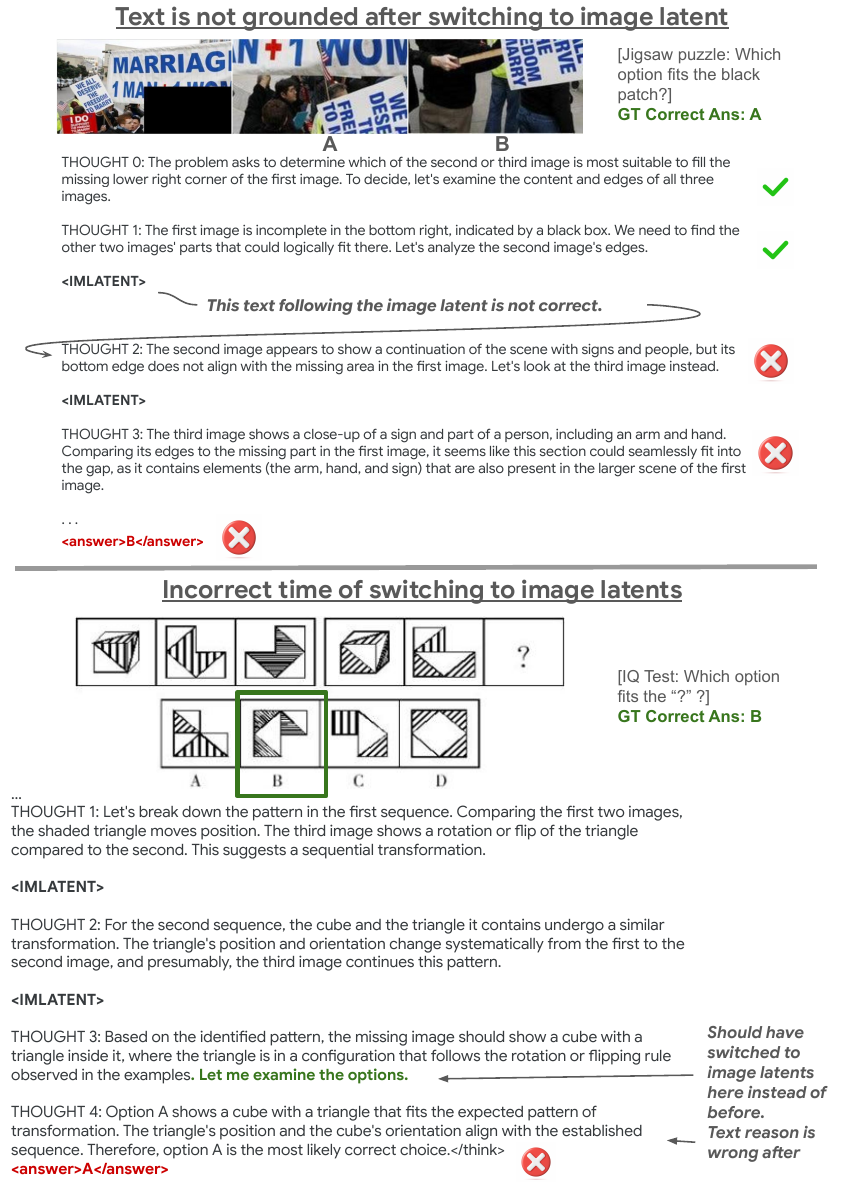}
\caption{Some qualitative examples of failures we observe with trying the existing approach of using explicit switching between image and text thoughts, motivating the need for our modality-agnostic \name{}. Upper half: the image thought may not be enough for the text following it to be accurately grounded. The model correctly identifies that it needs to think in the image latent, but the text after is inaccurate. With our \name{}, the model can free-form decide how many latents to allocate to the image since it doesn't need to explicitly switch back to text. Bottom half: Sometimes, the model cannot correctly identify when to switch to image latents. It switches at suboptimal times and doesn't switch to image latent even after identifying it needs to now compare the visual options (green bolded text). Our modality-agnostic \name{} removes the need to explicitly switch between modalities. }
\label{fig:qual_im_txt_interleave}
\end{figure*}

\subsection{Training Details}

We train using Deepspeed~\cite{deepspeed_kdd2020} stage 2 on 8 H100 GPUs. 
For the fine-tuned versions, we use a batch size of 1.
We use a max gradient norm of 5 and use float bf16 precision for all our experiments, with a learning rate of $1\textrm{e}{-6}$. 
We freeze the vision encoder since we see minimal differences with training the visual encoder in our early experiments. 
All other parts of the model are fully fine-tuned. 

The GRPO variants are trained using Deepspeed stage 3 to accommodate the significant increase in memory requirements. 
For the GRPO we use a reference model which is a frozen version of the same model. 
We use a group size of 2 (we run out of memory with a higher group size). 
We use a gradient accumulation size of 32 with 8 GPUs (data parallel). We use a beta (weighing the KL divergence) of 0.04. 
These hyperparameters follow the standard settings in the Video-R1 codebase \cite{feng2025videor1}. 
We do GRPO for 200 steps - hence, 50K iterations ($200 \times 32 \times 8$). 

\xhdr{Data mix used for each stage.}
For stage 1, our goal is to primarily pretrain the \name{} to hold intermediate reasoning information. Hence, we primarily use the Video-R1 and Zebra-COT datasets which include reasoning traces. Specifically, we use a mixture of $40\%$ Video-R1, $40\%$ Zebra-COT, and $20\%$ SAT for 200K iterations. 
In Stage 2 and 3, we now want the model to optimize the reasoning chains to data primarily not seen in the pretraining SFT mixture. Hence, we use a higher proportion of SAT and still include some pretraining data to prevent domain forgetting following \cite{ray2024sat}. Specifically, we use a mixture of $60\%$ SAT, and $20\%$ each from Video-R1 and Zebra-COT for another 200K iterations. 

For the baselines, we maintain the number of iteration steps to ensure fair comparison. 
For DirAns FT baseline - we also first train for 192K iterations (24K steps with data parallel on 8 GPUs) on  $40\%$ Video-R1, $40\%$ Zebra-COT, and $20\%$ SAT and then on $60\%$ SAT, and $20\%$ each from Video-R1 and Zebra-COT for another 192K iterations. We also tried training all in one stage for longer using $60\%$ SAT, and $20\%$ each from Video-R1 and Zebra-COT for 384K iterations and found the former to be a stroner baseline.
For the TextCOT baseline, we first train for 192K iterations on  $40\%$ Video-R1 with text reasoning traces, $40\%$ Zebra-COT answers only, and $20\%$ SAT and then do GRPO for 50K steps on $60\%$ SAT, and $20\%$ each from Video-R1 (answers only) and Zebra-COT (answers only). 
We also tried the baseline of only first training on the Video-R1 since it is the only dataset with text reasoning traces before doing GRPO. However, we found the base model trained first with the entire mix to be a stronger starting point (higher benchmark numbers) for the GRPO.

\subsubsection*{Prompting Strategies \& Evaluation Settings}

To strictly evaluate the reasoning capabilities of different model configurations, we employ three distinct prompting strategies. The specific template chosen determines whether the model relies on explicit textual generation, latent computation, or direct pattern matching. Additionally, a task-specific constraint (e.g., \textit{"Please provide only the single option letter"} or \textit{"Please provide a numeric answer"}) is appended to all templates based on the question type.

\vspace{0.5em}
\noindent \textbf{1. Text-Reasoning Baseline (Video-R1).} \\
To reproduce text-based reasoning baselines, we utilize the template established in prior work~\cite{feng2025videor1, yang2025mirage, luo2025thinkingdriftsevidentialgrounding}. 

\begin{promptbox}
\{Question\} \\
Please think about this question as if you were a human pondering deeply. Engage in an internal dialogue using expressions such as 'let me think', 'wait', 'Hmm', 'oh, I see', 'let's break it down', etc, or other natural language thought expressions. It's encouraged to include self-reflection or verification in the reasoning process. Provide your detailed reasoning between the $<$think$>$ $<$/think$>$ tags, and then give your final answer between the $<$answer$>$ $<$/answer$>$ tags.
\end{promptbox}

\vspace{0.5em}
\noindent \textbf{2. Image-text interleaved and \name{} (Ours).} \\
For models based on our proposed \name{}, we employ a similar template. Since we do not explicitly verbalize the thoughts, we avoid phrases that encourage expressions like ``Hmm", ``Oh, I see' etc.  

\begin{promptbox}
\{Question\} \\
Please think about this question deeply. It's encouraged to include self-reflection or verification in the reasoning process. Provide your final answer between the $<$answer$>$ $<$/answer$>$ tags.
\end{promptbox}

For image-text interleaved thoughts, we use Zebra-CoT~\cite{li2025zebra} dataset. Each image is first encoded by the Qwen-VL~\cite{Qwen-VL} encoder, then all tokens are average-pooled into a single image feature vector. While allocating more latents to each image can likely help at the cost of higher compute, we leave such an ablation to future work. We use a discrete token mapped to the image feature, rather than using continuous embeddings for two reasons. First, we observed a slight degradation in performance.  Second, continuous embeddings are slower due to recurrence loops during inference and training.   
During training, we use an explicit token \texttt{<im start>} followed by the \texttt{<latent>} token. The output of the \texttt{<latent>} is supervised to be the image feature. 

For the \name{} approach, we simply append $k$ \mulltt{} after the question prompt, providing the model with additional tokens as a multimodal thinking scratchpad. 

\vspace{0.5em}
\noindent \textbf{3. Direct Answering (SFT \& Base Model).} \\
Finally, to evaluate the base model (zero-shot) and the standard Supervised Fine-Tuning (SFT) baseline, we use a \textit{Direct} template. This is the most barebones setting to assess the model's ability to map inputs directly to answers without intermediate computation steps.
\begin{promptbox}
\{Question\} \\
Provide your final answer between the $<$answer$>$ $<$/answer$>$ tags.
\end{promptbox}

\noindent \textbf{Task-Specific Output Constraints.} \\
To ensure the generated answers can be reliably parsed and evaluated against ground-truth, we append a format constraint to every prompt during evaluation. These constraints vary by task definition (e.g., Multiple Choice vs. Regression) and dictate the expected content within the \texttt{<answer>} tags. The specific suffixes used are:

\begin{itemize} \setlength\itemsep{0.2em}
    \item \textbf{Multiple Choice:} 
    \textit{\texttt{"Please provide only the single option letter (e.g., A, B, C, D, etc.) within the $<$answer$>$ $<$/answer$>$ tags."}}
    
    \item \textbf{Numerical / Regression:} 
    \textit{\texttt{"Please provide the numerical value (e.g., 42 or 3.14) within the $<$answer$>$ $<$/answer$>$ tags."}}
    
    \item \textbf{OCR:} 
    \textit{\texttt{"Please transcribe text from the image/video clearly and provide your text answer within the $<$answer$>$ $<$/answer$>$ tags."}}
    
    \item \textbf{Free-form:} 
    \textit{\texttt{"Please provide your text answer within the $<$answer$>$ $<$/answer$>$ tags."}}
\end{itemize}

\noindent \textbf{Fuzzy Logic Used to Extract Answers} \\
To ensure robust evaluation, particularly when models produce verbose CoT outputs, we implement a fuzzy matching strategy to parse the final prediction as follows:

\begin{itemize}
    \item \textbf{Pattern Extraction:} We first attempt to extract the answer by matching the prediction against a prioritized list of regular expression patterns. These include standard conversational markers and our enforced XML tags:
    \begin{itemize}
        \item \texttt{r"<answer>\textbackslash s*(.*?)\textbackslash s*</answer>"}
        \item \texttt{r"(?:the answer is|the correct answer is|final answer:)\textbackslash s+([a-zA-Z0-9]+)"}
        \item \texttt{r"I counted a total of\textbackslash s+(\textbackslash d+)"}
        \item \texttt{r"count is\textbackslash s+(\textbackslash d+)"}
    \end{itemize}

    \item \textbf{Validation \& Numerical Extraction:} If a match is found, we check if the extracted content is a valid multiple-choice option (single letter A--Z). If the extracted text is not a letter, we attempt to parse it as a numerical value (integer or float) for regression or counting tasks.
    
    \item \textbf{Fallback Strategy:} In cases where the extraction fails or yields no numerical/categorical value, we perform a final fallback search for the last isolated capital letter (e.g., \texttt{r"\textbackslash b[A-Z]\textbackslash b"}) appearing in the raw text, assuming it represents the final selection.
\end{itemize}

Empirically,  we observe that the models usually adhere to the formatting instructions, with the vast majority of valid answers being successfully parsed directly from the \texttt{<answer>} tags via the first pattern.

\subsection{Further Analysis and Ablations}
In this section, we examine the shortcomings of modality switching, contrasting the inductive biases of interleaved image-text generation against our proposed joint reasoning approach.  We identify specific issues, including ungrounded modality transitions and suboptimal scheduling, which hinder the performance of explicit switching baselines. Finally, we offer a preliminary analysis of the latent dynamics within our \name{}.

\subsubsection*{The failure points with interleaving image-text}

To investigate the feasibility of explicit modality switching for visual reasoning, we conducted a pilot study using a baseline model trained to interleave textual reasoning traces with discrete image latent tokens. We hypothesize that distinct processing steps for visual and textual thinking should allow for better interpretability and performance. However, our experiments reveal critical limitations in this paradigm.

\xhdr{Bias Towards Textual Reasoning.}
We observed a strong inductive bias in the model to rely solely on textual reasoning. When trained simply to solve problems, the model rarely invoked image latents at test time, preferring to reason exclusively via text. 
To address this, we introduced specific system prompts during training: one explicitly instructing the model to ``think only using text'' when the dataset included text-based reasoning traces and another to ``think using both text and images'' for the image-text interleaved training data. 

Now, at test-time, we tried two approaches - one where we let the model decide by simply including the question without instructions for how to reason, and one where we append ``think using both text and images'' after the question to \textit{force} the model to think using images. The results are shown in Table \ref{tab:im_text_interleave}.

We observe two issues during inference:
\begin{itemize}
    \item \textbf{Bias towards text-only reasoning:} When given the autonomy to choose its reasoning modality (i.e., without a forcing prompt), the model mostly reverts to text-only reasoning, and rarely switches to image latent thinking.
    \item \textbf{Forcing image thoughts degrades performance:} When we explicitly forced the utilization of image latents via the prompt, we observed a degradation in downstream task performance compared to the text-only baseline, as shown in Table~\ref{tab:im_text_interleave}.
\end{itemize}

\xhdr{Performance Degradation with Forcing image Thoughts.} 
Comparing the \textit{Img-Txt (free-form)} baseline with the \textit{Img-Txt (force im thought)} setting reveals that prompting the model to use image thoughts leads to a reduction in overall performance. The average score across all benchmarks drops from $50.5\%$ to $47.3\%$. 
Despite the aggregate decline, the forced image thought setting yields marginal but distinct improvements on specific tasks that inherently require visual simulation or geometric imagination- such as Jigsaw ($68.7 \rightarrow 69.3$) and Relative Depth (RelDep) ($69 \rightarrow 73$) -  tasks that may  require mentally manipulating visual patches to solve spatial puzzles, or visualizing depth maps or 3D structures from the 2D RGB inputs.
However, while specific cases may benefit, we see an overall decline in performance compared to letting the model decide. 
These results suggest that while explicit image thoughts intuitively should contain valuable signals, the mechanism of explicit switching may be too brittle for general reasoning.

\xhdr{Qualitative Analysis of Switching Failures.}
To investigate performance degradation during explicit switching, we also qualitatively analyzed the model's generated traces. We identified two prominent patterns in the failure cases that we show in Figure~\ref{fig:qual_im_txt_interleave}:

\begin{enumerate}
    \item \textbf{Ungrounded Modality Transitions:} As shown in the top row of Figure~\ref{fig:qual_im_txt_interleave}, even when the model successfully generates an image latent (intended to process visual cues like fitting in the puzzle piece), the subsequent textual reasoning is often not properly grounded in that latent. In this example, The model correctly identifies that it needs to analyze the image choice's edges and hence switches to the image thoughts. However, the text following it is incorrect. This suggests that the image latent did not contain enough information for the text thought to be properly grounded.
    
    \item \textbf{Suboptimal Switching Scheduling:} The bottom row of Figure~\ref{fig:qual_im_txt_interleave} demonstrates that the model struggles to identify the optimal moment to switch modalities. In this example, the model should have switched to latent after identifying the need to examine the options. 
\end{enumerate}

These findings suggest that rigid, explicit switching between modalities may be difficult and require more data. This limitation motivates our proposed approach: \emph{modality-agnostic thinking tokens} that can jointly reason in a continuous space without explicit switching.

This motivates our modality agnostic \name{}. \name{} offers a simpler alternative where we do not need to switch between modalities, but rather think using latents trained to hold both visual and textual information jointly.
Our proposed \textit{\name{}} approach achieves the highest performance overall (Avg $53.9\%$). We posit that by enabling joint reasoning in a shared space, \name{} capture the benefits of visual imagination (e.g., high performance on RelDep and Jigsaw) without suffering the grounding losses associated with explicit modality switching. 
However, more interpretability studies will need to be done as future work to rigorously validate the function of \name{}. 

\subsubsection*{A preliminary attempt to interpret latents}
While our primary focus has been on performance, analyzing the semantic content of the \mulltt{} latents presents a compelling avenue for future work. 
We conducted a preliminary analysis of the Euclidean distance between subsequent latents to see if latents encode different meanings. For \name{} without a warm-up of image-text traces, the Euclidean distance between latents tapers off after a few steps before it converges, suggesting they become similar and add no new information. Our image-text warmed up latents maintains distances between steps. This indicates that the model avoids representational collapse and that each latent continues to contribute information (albeit useful or not). 
However, more analysis needs to be done by decoding or mapping the latents to actual reasoning images and closely controlled counterfactual images to understand and interpret the ``meaning'' of the \name{}. With the absence of such data, we leave this analysis to future work. Our focus in the paper is to provide a more accurate, simple, and fast alternative to text reasoning or explicit image thoughts that require 100's of tokens as reasoning.


 



\end{document}